\documentclass{IEEEtrans4PSCC}
\usepackage{cite}
\usepackage{color}
\usepackage{soul}
\ifCLASSINFOpdf
   \usepackage[pdftex]{graphicx}
\else
   \usepackage[dvips]{graphicx}
\fi
\usepackage[cmex10]{amsmath}
\makeatletter
\let\old@ps@headings\ps@headings
\let\old@ps@IEEEtitlepagestyle\ps@IEEEtitlepagestyle
\def\psccfooter#1{%
    \def\ps@headings{%
        \old@ps@headings%
        \def\@oddfoot{\strut\hfill#1\hfill\strut}%
        \def\@evenfoot{\strut\hfill#1\hfill\strut}%
    }%
    \def\ps@IEEEtitlepagestyle{%
        \old@ps@IEEEtitlepagestyle%
        \def\@oddfoot{\strut\hfill#1\hfill\strut}%
        \def\@evenfoot{\strut\hfill#1\hfill\strut}%
    }%
    \ps@headings%
}
\makeatother

\psccfooter{%
        \parbox{\textwidth}{\hrulefill \\ \small{23rd Power Systems Computation Conference} \hfill \begin{minipage}{0.2\textwidth}\centering \vspace*{4pt} \includegraphics[scale=0.06]{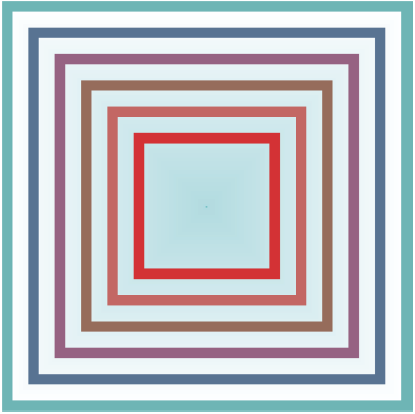}\\\small{PSCC 2024} \end{minipage} \hfill \small{Paris, France --- June 4 -- June 7, 2024}}%
}
\def\tablename{Table}
\usepackage{amssymb}
\usepackage{mathtools, nccmath}
\usepackage{amsmath}
\usepackage{import}
\usepackage{hyperref}
\usepackage{float}
\usepackage{multirow}
\usepackage{booktabs}
\usepackage{caption}
\usepackage{graphicx}
\usepackage{subcaption}
\usepackage{mwe}
\usepackage{algorithmic}
\usepackage{algorithm}
\usepackage{comment}
\begin{document}
\title{Learning a Reward Function for User-Preferred Appliance Scheduling}



\author{
\IEEEauthorblockN{Nikolina Čović}
\IEEEauthorblockA{University of Zagreb\\Faculty of Electrical\\Engineering and Computing\\Zagreb, Croatia\\
Email: nikolina.covic@fer.hr}
\and
\IEEEauthorblockN{Jochen L. Cremer}
\IEEEauthorblockA{Delft University of Technology\\Faculty of Electrical Engineering,\\ Mathematics, and Computer Science\\Delft, The Netherlands\\
Email: j.l.cremer@tudelft.nl}
\and
\IEEEauthorblockN{Hrvoje Pandžić}
\IEEEauthorblockA{University of Zagreb\\Faculty of Electrical\\Engineering and Computing\\Zagreb, Croatia\\
Email: hrvoje.pandzic@fer.hr}
}

\maketitle

\begin{abstract}
Accelerated development of demand response service provision by the residential sector is crucial for reducing carbon-emissions in the power sector. Along with the infrastructure advancement, encouraging the end users to participate is crucial. End users highly value their privacy and control, and want to be included in the service design and decision-making process when creating the daily appliance operation schedules. Furthermore, unless they are financially or environmentally motivated, they are generally not prepared to sacrifice their comfort to help balance the power system. In this paper, we present an inverse-reinforcement-learning-based model that helps create the end users' daily appliance schedules without asking them to explicitly state their needs and wishes. By using their past consumption data, the end consumers will implicitly participate in the creation of those decisions and will thus be motivated to continue participating in the provision of demand response services.
\end{abstract}


\thanksto{This work was supported by the Croatian Science Foundation under project Active NeIghborhoods energy Markets pArTicipatION—ANIMATION (IP-2019-04-09164) and the WeForming project that has received funding from the European Union’s Horizon Europe Programme under the Grant Agreement No. 101123556. Employment of Nikolina Čović is fully funded by the Croatian Science Foundation under project DOK-2020-01-3911. }

\section{Introduction}
To accommodate the growth of electricity demand within the existing power grid, demand and generation must be locally balanced as much as possible. Energy communities allow their participants to respond with demand and generation adjustments to maintain the local energy balance. For example, a community participant (user) with an appliance such as a washing machine may be willing to shift operating the machine from demand peaks to times of higher power generation or to times when their neighbor's battery is fully charged. However, participation in these community concepts often lacks the right incentives, which represents an obstacle to forming energy communities. 

The reasons for the insufficient deployment of energy communities are structural and motivational \cite{TNO2022}. 
Structurally, individuals may find technology, such as smart appliances, too expensive, insecure, or unaccommodating. The additional investments required to control their appliances may be presented as an unnecessary financial burden, while potentially insufficiently secure information sharing may feel threatening to some users. These new technologies may seem quite menacing since most users tend to appreciate their autonomy and ownership, which is why relinquishing control of appliances could prove to be an extremely difficult step. Therefore, they decide not to participate. Motivationally, possibly impaired comfort of the user is identified as one of the key reasons for not participating in the provision of the service. Individuals may be hesitant to participate or lack persistence due to expected discomforts despite the community benefits to the environment and their economics. Furthermore, human decision-making hesitations favor the status quo of not participating. In past research, the structural-technological reasons have been studied significantly, demonstrating that the financial benefits scale (economy of scale) and control algorithms may exist while managing users' privacy and autonomy. However, the motivational reasons were not studied adequately in these algorithms. 
In response, this paper proposes a control algorithm that directly interacts with (and learns from) the user, resolving issues of hesitations and persistence by optimally balancing the expected discomforts with their motivations, whether they are environmental or economic.


Household device management usually defines the objective function as a weighted sum of the predefined energy cost and the user (dis)satisfaction functions \cite{7438871, ali2022, nowbandegani2022, alzahrani2023, Til23}. Similarly, most research papers arbitrarily define weights and analyze the sensitivity of the model to these weights, for example, \cite{xie2018}. 
Accordingly, the authors in \cite{chai2016} present a multi-objective optimization model for the optimal operation scheduling of shiftable household appliances. By using an iterative learning technique, they are trying to determine the weight value that will enable an optimal trade-off between achieving the minimum energy cost and the minimum user dissatisfaction, which are often in conflict. As a result, consumption costs are reduced, and the user's comfort remains intact. However, the presented model as the all available literature, to the best of our knowledge, assumes that the user's discomfort can be modeled by a predefined mathematical function. Due to each person having different preferences, a single discomfort function probably will not fit all users, so a proper control algorithm should be able to detect an appropriate discomfort function for each user. It might even be difficult to fit a single user throughout an indefinite time period. On shorter terms, people's preferences might change from morning until evening, just as they can change on a daily basis. In the longer course of time, a person may change their lifestyle and would need the control algorithm to recognize the change. 

Another issue of presented methods that illustrates itself precisely in the model's inability to adapt to a changed environment stems from utilizing methods that require knowledge of all data beforehand (e.g. deterministic optimization). Reinforcement learning, on the other hand, in addition to its ability to learn offline, allows learning through interactions with the environment online, leaving room to improve the adaptability of the model \cite{nagabandi2019}. This technique adds to the versatility of the model by making decisions, unlike the other mentioned approaches, solely based on the \textit{interactions} with the environment, without actually requiring its exact mathematical model. This way, the developed algorithm can be used in completely different environments --- in households with different appliances or the same appliances of different brands. Nevertheless, although reinforcement learning solves the issue of adaptability of the model to different households and to different/changed human habits, the available literature, e.g. \cite{8919976, 9085925, yu2021, zou2020, gupta2021}, still uses fixed and predefined reward functions, which leaves the problem of unknown basic model of discomfort still unsolved. 



A promising field of inverse reinforcement learning (IRL), whose foundations were laid by Ng and Russell in \cite{Ng2000}, gives a solution to the remaining matter by learning the reward function from data on previous user behavior. Particularly, allowing the user's data to speak for itself, while making no assumptions about the user's preferences beforehand, is the greatest benefit of this approach. Numerous works deal with learning the reward function in the field of robotics. \cite{sadigh2017} presents a method for learning the reward function based on the user's preferences expressed through their direct feedback and show successful results in the semi-autonomous driving example. Similarly, in \cite{Biyik2019} the authors propose reward learning from feedback that provides the most information to achieve the best solution. Once again, the authors simulate the theoretical considerations on the example of autonomous driving and two other robotic environments. Contrarily, no work has been found on reward function learning in power systems per our knowledge. 

A household owning power-curtailable and time-shiftable appliances provides demand response services through which it can obtain financial benefits, while also potentially impair its comfort. This paper attempts to find the underlying reward function that guides the user's daily decision-making process of household appliance scheduling solely based on the examples of the user's past demand response provisions using inverse reinforcement learning, without explicitly modeling the user's habits and wishes. It tackles the following issues present in the current demand response provision literature:
\begin{enumerate}
\item Unknown reward/objective function that guides the behavior;
\item Adaptability of the model to short-/long-term changes in human behaviour;
\item Adaptability of the model to different users.
\end{enumerate}

\section{Demand Response as a Markov Decision Process}
\label{sec:meth} 

The problem of demand response can be modeled in a simplified manner by a Markov Decision Process. Markov decision process is a stochastic model in which the current state depends only on the one that preceded it. Such a model is defined by a set of states $\mathcal{S}$, a set of actions $\mathcal{A}$, the probability of transitioning from state $s$ to state $s'$ if an action $a$ is taken ($P_{s,a}^{s'}$), the reward $r$ in state $s$, and the discount factor $\gamma$. The agent in a certain state performs an action $a$, moves to state $s'$ with probability $P_{s,a}^{s'}$, and receives a reward $r$ for this action. A household providing demand response services can have their appliances clustered in three different categories:
\begin{enumerate}
    \item time-shiftable (washing machine, dishwasher, dryer, electric vehicle, etc.)
    \item power-curtailable (AC, electric heater, etc.)
    \item non-shiftable, 
\end{enumerate}
and based on their operation requirements, state and action spaces need to be determined. Time-shiftable devices can be shifted in time, however, once turned on, they operate at a constant power and cannot be turned off. On the other hand, power-curtailable devices can change their power output at any given time. The last category, non-shiftable devices, should never have their power output or time schedule altered and should always have their demand fully satisfied. The state space of time-shiftable devices consists of time periods in which the operation of the appliance was requested but delayed for demand reponse purposes, while in the case of power-curtailable and non-shiftable appliance it contains information about their demand at a given time period. The action space, on the other hand, is modeled as discrete reductions in \textit{total} consumption\footnote{This paper assumes demand response for congestion management to simplify the problem and reduce the action state space.} to multiples of a tenth of total household demand at a given time period, for example to 20\% or 60\% of total demand:

\begin{equation}
    \mathcal{A} = \{0, \;0.1 \;D, \;0.2\; D, \;...\;, \;D\}. \notag
\end{equation}
Later on, after the action is decided upon and, therefore, the total consumption in the given time period is determined as well, an optimization problem is run to distribute the total consumption per individual appliance. 

Reward function in a demand response problem is usually formulated in two parts --- on one side financial earnings from providing the service, while on the other, comfort disruption cost. Financial earnings are calculated as the price offered for service provision ($c$) multiplied by the change in consumption a user made as compared to the assumed baseline consumption\footnote{The baseline consumption of a single hour of the day is calculated as the average consumption in that hour for the last 10 days.} ($b$) it should have had without providing demand response. If demand response is used for congestion management purposes, only the reduction of consumption should be financially rewarded, as given in \eqref{eq. reww}. Comfort disruption, or discomfort ($\psi$), however, is commonly modeled by a quadratic function or an absolute value of the deviation from the previously set baseline value (e.g. desired room temperature). Nonetheless, since comfort experience can vary among different users, different times of the day or year, etc., it is not possible to define a single comfort function that works for all purposes.

\begin{equation}
    r = c \cdot \max{\left[0, (b-p)\right]} - \psi
    \label{eq. reww}
\end{equation}


Thus, at each hour of the day, the agent receives data on the household demand and makes decisions (chooses actions) on the amount of flexibility provided in the form of reduced consumption in order to achieve the highest reward. It is necessary to find such a sequence of actions $\{a_1,...,a_k\}$ that the user's total reward is ultimately the highest possible. The function that maps state $s$ to action $a$ is called a policy $\pi$, and one wants to find the one that is optimal. Goodness of the policy is most often determined by the value function $V(s)$, which gives the expected value of the total reward if we follow policy $\pi$ from state $s$, given by the Bellman eq. \eqref{eq. valuefunc}.

\begin{equation}
    V^{\pi}(s)= r(s) + \gamma \sum_{s'} P_{s,\pi(s)}(s')V^{\pi}(s')
    \label{eq. valuefunc}
\end{equation}

Furthermore, it is possible to define the value brought to the agent by doing action $a$ in state $s$ --- $Q(s,a)$, as in \eqref{eq. qualityfunc}.

\begin{equation}
Q^{\pi}(s,a)= r(s) + \gamma \sum_{s'} P_{s,a}(s')V^{\pi}(s')
    \label{eq. qualityfunc}
\end{equation}

By comparing the Q-values of different actions one can determine which action is better.

\section{Learning User Preference with IRL}
\label{sec:irl}
The scientific community has been working on developing algorithms for determining optimal $\pi$, which directly depends on the definition of the reward function. However, it is often impossible, or at least very difficult, to mathematically precisely determine the reward function for which the optimal behavior of the agent will then be sought. Ng and Russell in \cite{Ng2000} approach the problem from another angle. Assuming the knowledge of the optimal policy either through knowing the environment or the available trajectory samples, they formulated the problem of discovering the underlying reward function.

Starting from the fact that \eqref{eq. valuefunc} can be rewritten in a vector notation as $\textbf{V}^{\pi} = \textbf{R} + \gamma\textbf{P}_{a_{1}}\textbf{V}^{\pi}$ and $\pi(s) \equiv a_{1}$, then

\begin{equation}
\textbf{V}^{\pi}  = \left( \textbf{I} - \gamma \textbf{P}_{a_{1}}\right)^{-1} \textbf{R}
    \label{eq. formula5}
\end{equation}
and the reward must satisfy the following inequality:

\begin{equation}
\left( \textbf{P}_{a_{1}} - \textbf{P}_{a}\right) \left( \textbf{I} - \gamma \textbf{P}_{a_{1}}\right)^{-1} \textbf{R} \succeq 0
    \label{eq. formula4}
\end{equation}

Many rewards can satisfy this formula (including the one equal to 0), which is why the authors introduce additional conditions which the reward must satisfy in order to be considered appropriate: the reward must be such that any deviation from the optimal policy (choosing a suboptimal action) brings a large reduction to the value function (penalty). Furthermore, the reward must be as small as possible since it is assumed that it is better to have as simple rewards as possible. Therefore, the Q-value of the optimal action must be greater than all the other actions. By stating that condition as the objective function \eqref{eq. formula6}
\begin{equation}
\sum_{s \in \mathcal{S}} \left( Q^{\pi}(s,a_{1}) - \max_{a \in \mathcal{A} \setminus a_{1}}Q^{\pi}(s,a)\right)
    \label{eq. formula6}
\end{equation}

 and obeying constraint \eqref{eq. formula4}, the optimization problem can be formulated: 

\begin{gather}
\max \sum_{i=1}^{N} \min_{a \in \mathcal{A} \setminus a_{1}} \left[ \left (\textbf{P}_{a_{1}}(i) - \textbf{P}_{a}(i)\right) \left( \textbf{I} - \gamma \textbf{P}_{a_{1}}\right)^{-1}\textbf{R}\right] - \notag\\  \lambda ||\textbf{R}||_1 \notag\\
\hspace*{-1cm} \mathrm{s.t.} \left(\textbf{P}_{a_{1}} - \textbf{P}_{a}\right) \left( \textbf{I} - \gamma \textbf{P}_{a_{1}}\right)^{-1}\textbf{R} \succeq 0, \;\; \forall a \in \mathcal{A} \setminus \pi(s) \notag \\
|\textbf{R}_i| \le R_{\mathrm{max}},\;\; i = 1,...,N
    \label{eq. formula7}
\end{gather}
where $i$ denotes one of the $N$ states in state space $\mathcal{S}$, $\lambda$ represents an adjustable penalty that balances two parts of the objective function and $R_{\mathrm{max}}$ stands for the theoretical maximum of the reward value.

In order to further limit the search space of the reward function, the following linear approximation of reward function is assumed: 

\begin{equation}
R(s) = \alpha_1 \phi_1 (s) + \alpha_2 \phi_2 (s) + ... + \alpha_d \phi_d (s)
    \label{eq. formula8}
\end{equation}

Furthermore, most problems in the real world, including demand response, have unknown environments which is why it is only possible to access the state of the environment at a particular time period. For such problems, different state samples or trajectories are collected from which policies can be determined, which are then used to calculate the expected value of the value function as

\begin{gather}
    \hat{V}^{\pi} (s_0) = \alpha_1 \hat{V}^{\pi}_1 (s_0) + \cdot\cdot\cdot + \alpha_{d} \hat{V}^{\pi}_d (s_0), \notag \\
    \hspace*{-1cm}\mathrm{where}  \;\;\hat{V}^{\pi}_i (s_0) = \phi_i (s_0) + \gamma\phi_i (s_1) + \gamma^{2}\phi_i (s_2) + \cdot\cdot\cdot
    \label{eq. formula11}
\end{gather}

Finally, the optimization problem transforms into:

\begin{gather}
    \max_{\alpha_i} \sum_{s_0 \in \mathrm{S}_0} \sum_{j=1}^{k} \left (\hat{V}^{\pi^{*}}(s_0) - \hat{V}^{\pi_{j}}(s_0) \right )  \notag\\
    \hspace*{-1cm} \mathrm{s.t.} \,\;\;\;\: |\alpha_i| \le 1, \,\;i=1,...,d 
    \label{eq. formula13}
\end{gather}
and is solved according to Algorithm \ref{alg: pseud}. The requirements of the algorithm are the optimal policy $\pi^*$, which is assumed to be known for the demand response case through the user's past consumption data, and a randomly chosen policy $\pi_1$\footnote{An example of the policy $\pi_1$ is to choose a random action $a$ at a state $s$}. For a more detailed mathematical formulation, please refer to \cite{Ng2000}.
\begin{algorithm}
    \caption{IRL algorithm}
    \begin{algorithmic} 
    \REQUIRE $\Pi = \{\pi^*, \pi_1\}, \; V_i^{\pi_j}(s_0), \; k=2$
    \WHILE{$k<N$}
    \STATE $\alpha_i \leftarrow$ optimization problem in \eqref{eq. formula13} for policies in $\Pi$
    \STATE $\pi_k \leftarrow$ train the agent with $\alpha_i$ and eq. \eqref{eq. formula8}
    \STATE add $\pi_k$ to $\Pi$
    \IF{performance is satisfying}
    \STATE  break
    \ENDIF
    \STATE $k \leftarrow k + 1$ 
    \ENDWHILE
    \end{algorithmic}
    \label{alg: pseud}
\end{algorithm}

\section{Input Data and Model Setup}
The Pecan Street data set, which consists of data on consumption per individual household appliance for 25 households, was used \cite{pecandata}. Out of time-shiftable appliances, the data set contains consumption data for an electric vehicle, a washing machine, a dishwasher and a dryer. Power-curtailable representative is the air conditioning, while the rest of the data falls under the category of non-shiftable devices. Depending on the user, various combinations of household appliances per household are possible (some users do not own an electric vehicle, some do not have a dishwasher, etc.) Data is available from the beginning of April 2018 until beginning of November in the same year. The training set covers the periods from April 1st until July 1st and from August 2nd until November 1st. The period in the middle is used for testing purposes. 

\label{sec:inp}
Still, the available data show household consumption without demand response service provision. Due to the lack of consumption data of households that actually provide demand response service, which is required as an input (serves as the desired behavior) for the IRL algorithm from section \ref{sec:irl}, demand response provision is foremost simulated by a reinforcement learning (RL) model. So, for the algorithm verifying purposes, it is assumed that the household with the consumption from the available data set is providing demand response under the following reward function \eqref{eq. reww2}:
\begin{equation}
    r = c \cdot \max{\left[0, (b-p)\right]} - w_{ac} \cdot (\Delta p_{ac})^2 - \sum_{TS}w_m \cdot (t^d_m)^2
    \label{eq. reww2}
\end{equation}
where the discomfort is assumed to follow the quadratic function of air conditioning consumption deviation ($\Delta p_{ac}$) and quadratic function of time delay for each time-shiftable device ($t^d_m$). Coefficients $w_{ac}$ and $ w_m$ are measures of dissatisfaction --- the higher the value, the higher the inconvenience when the appliance schedule is changed. With this approach, the optimal policy $\pi^*$ under the reward with assumed discomfort modeling is obtained and is considered to be the historic behavior of the user who provided demand response services. Nonetheless, it is important to note that in the case of having a data set that contains consumption data of a household providing demand response, the previously described approach of generating data should be omitted and the available data should be used directly.

The basis functions $\phi_i(s)$ for approximating the reward function in eq. \eqref{eq. formula8} are domain-specific and should be carefully designed. Therefore, six different basis functions were assumed in this problem, which differed in the way the profit from the demand response service is calculated and in the functions that model the discomfort. More specifically, profit could be calculated as $\max{\left[0, (b-p)\right]}$ or just $(b-p)$ multiplied by price $c$. Discomfort could be neglected, considered just as the deviation in the AC consumption and time delay or by taking the square of deviation and time delays. Combining two ways of calculating the financial part of the reward function and three ways of calculating the discomfort part, results in six total combinations. Then optimizing coefficients $\alpha_i$ should find the appropriate reward function approximation that matches the desired behavior. 

The agent is trained on different number of episodes (1500, 2500 and 3500), depending on the case study, where one episode corresponds to a single day divided into 96 time steps of 15 minutes each. Deep Q Network (DQN) with 1 hidden layer of 32 units and a ReLU activation function after the first two layers is used. All three layers are linear. The input data to the network is the state of the household, which consists of the demand for power-curtailable and non-shiftable devices and open delays for time-shiftable devices (total of 6 data points), the unit price of providing flexibility, and the baseline demand. The output of network is the action, that is, the reduction in the total household consumption. To select the agent's action during the training process, an $\epsilon$-greedy strategy with a decreasing value of $\epsilon$ is implemented. The initial value of epsilon is 1.0, and with each episode, the value drops to 99.9\% of the previous one. Furthermore, during training, a soft update of the target network is used with the value of the update parameter $\tau = 0.001$. Among other hyperparameters important for the training, the batch size of 32, the discount rate when updating DQN network parameters of 0.9, the learning rate of 0.001, and the buffer size of 10,000 input data were selected. The end of the IRL algorithm is set to be reached after 10 iterations. 

\section{Case Studies}
In this section, three case studies are presented to obtain answers to the following queries:
\begin{enumerate}
\item whether the developed model can even succeed in reproducing the desired user behavior;
\item whether it is possible to find an approximate reward function that gives good enough results for different days / consumptions in a particular household;
\item whether the model will be able to properly function in households guided by a different reward function;
\end{enumerate}

All simulations were performed using Gurobi 9.1.2 and run on an AMD Ryzen 7 4700U CPU with 16 GB of RAM at 3733 MHz. {The code used in this paper is available in \cite{Cov23} and the data can be downloaded from \cite{pecandata}.
\subsection{Single Day}

To determine whether the presented model is at all capable of adequately approximating the reward function and recreating the desired behavior of the end user that provides a demand response service (obtained by following policy $\pi^*$ for the observed day, explained in section \ref{sec:inp}), the model was trained on two arbitrarily chosen representative days from the training set: one with few active controllable devices (April 2nd; trained on 1500 episodes) and one in which all controllable devices are running (August 31st; trained on 2500 episodes). Fig. \ref{fig:four graphs} shows the rewards and consumption changes made by the end user following the policy $\pi^*$ (desired, \textit{optimal} behavior; in blue) and the behavior the user learned guided by the approximate reward function (in green). In order to compare and validate the results, both policies (optimal and learned) were tested with the true reward function\footnote{This is only possible due to the lack of demand response consumption data and assumed reward function for generating policy $\pi^*$.} for both days. For both days, it can be observed that the reward converges to the optimal one after a few iterations. However, in the case of the day with lower usage of controllable devices, the agent's decisions obtained with the learned reward mirror the decisions obtained with the policy $\pi^*$, given in Fig. \ref{fig:changes92}. Similarly, the corresponding reward reaches the exact same value in the final iteration as with the policy $\pi^*$, as seen in Fig. \ref{fig:rews92}. Increasing the number of active devices in a day also makes it difficult to approximate the reward function, as seen in Figs. \ref{fig:changes243} and \ref{fig:rews243}. Although the agent's decisions guided by the learned policy were similar to the optimal ones, the accuracy considerably dropped compared to the case of April 2. This can be read more clearly in Table \ref{tb:metricsS} where, for the 31st August, the mean absolute error (MAE) was 0.12, versus 0.00 for the 2nd April. The same can be recognized with the remaining two metrics given in Table\footnote{MAE and MSE throughout this paper are in the range of $[0,1]$.}. Finally, the presented model can satisfactorily reproduce the desired behavior, but increasing the number of controllable devices makes the approximation of the reward more difficult.

\begin{figure*}
     \centering
     \begin{subfigure}[b]{0.475\textwidth}
         \centering
         \includegraphics[width=0.975\textwidth]{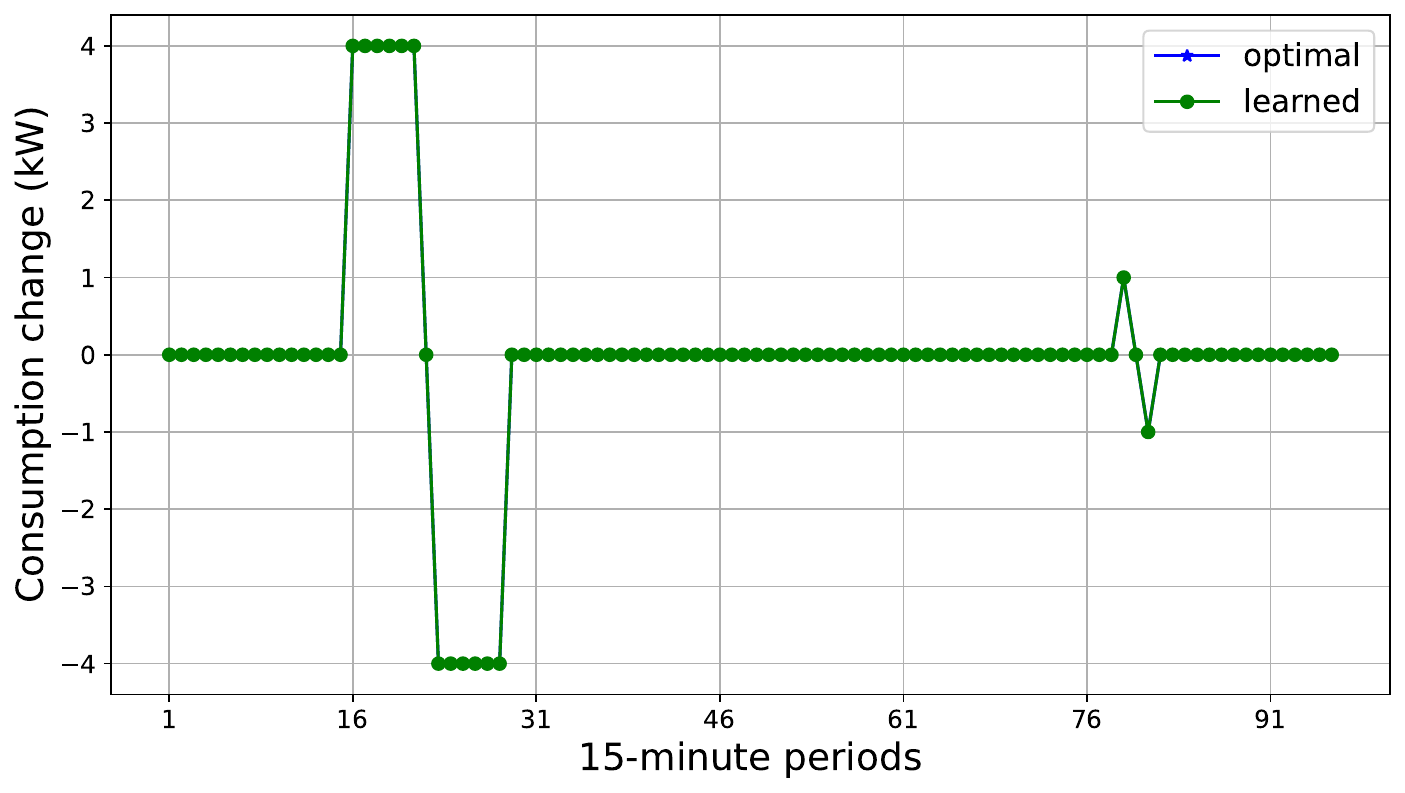}
         \caption{Changes in consumption on April 2nd}
         \label{fig:changes92}
     \end{subfigure}
     \hfill
     \begin{subfigure}[b]{0.475\textwidth}
         \centering
         \includegraphics[width=\textwidth]{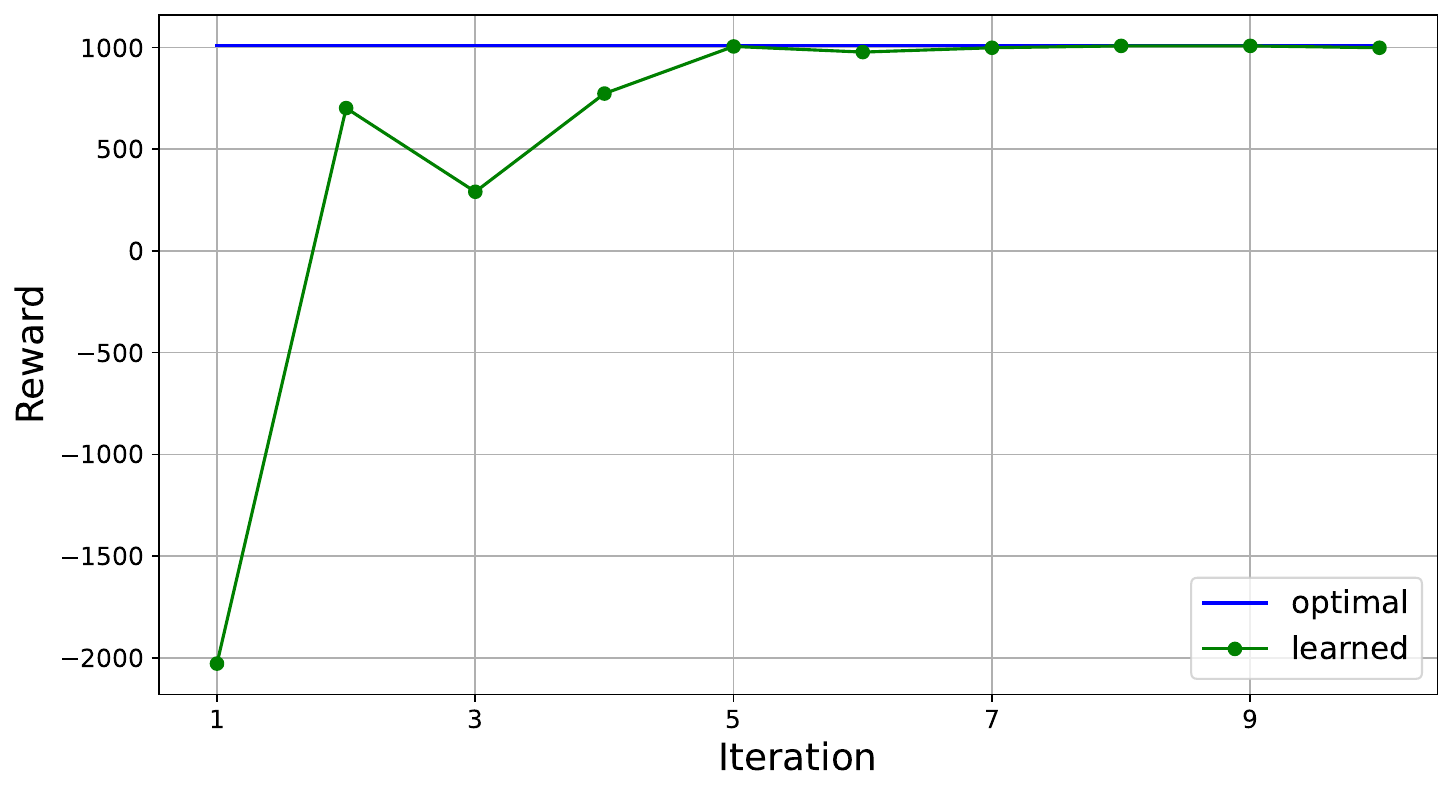}
         \caption{Obtained rewards on April 2nd}
         \label{fig:rews92}
     \end{subfigure}
     \vskip\baselineskip
      \begin{subfigure}[b]{0.475\textwidth}
         \centering
         \includegraphics[width=0.975\textwidth]{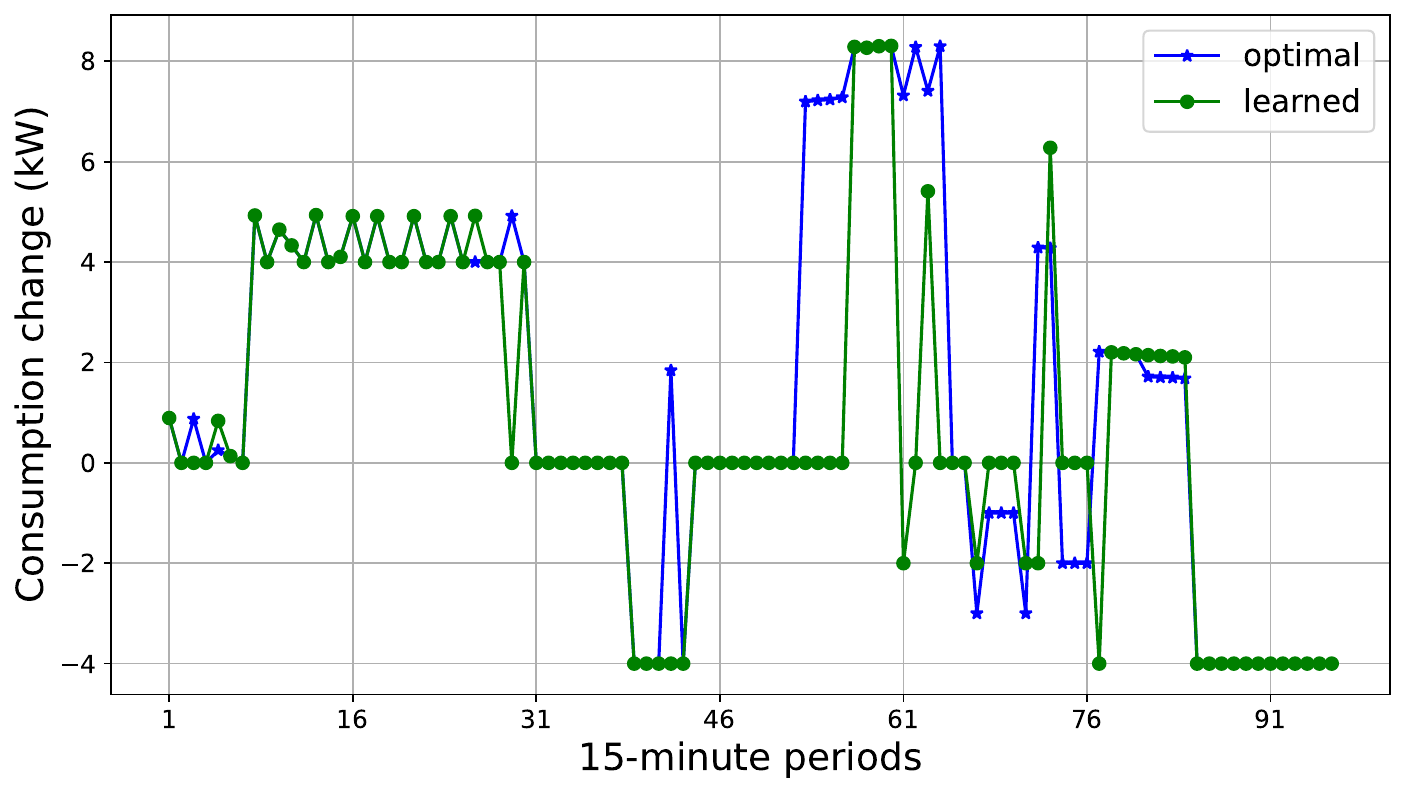}
         \caption{Changes in consumption on August 31st}
         \label{fig:changes243}
     \end{subfigure}
     \hfill
     \begin{subfigure}[b]{0.475\textwidth}
         \centering
         \includegraphics[width=\textwidth]{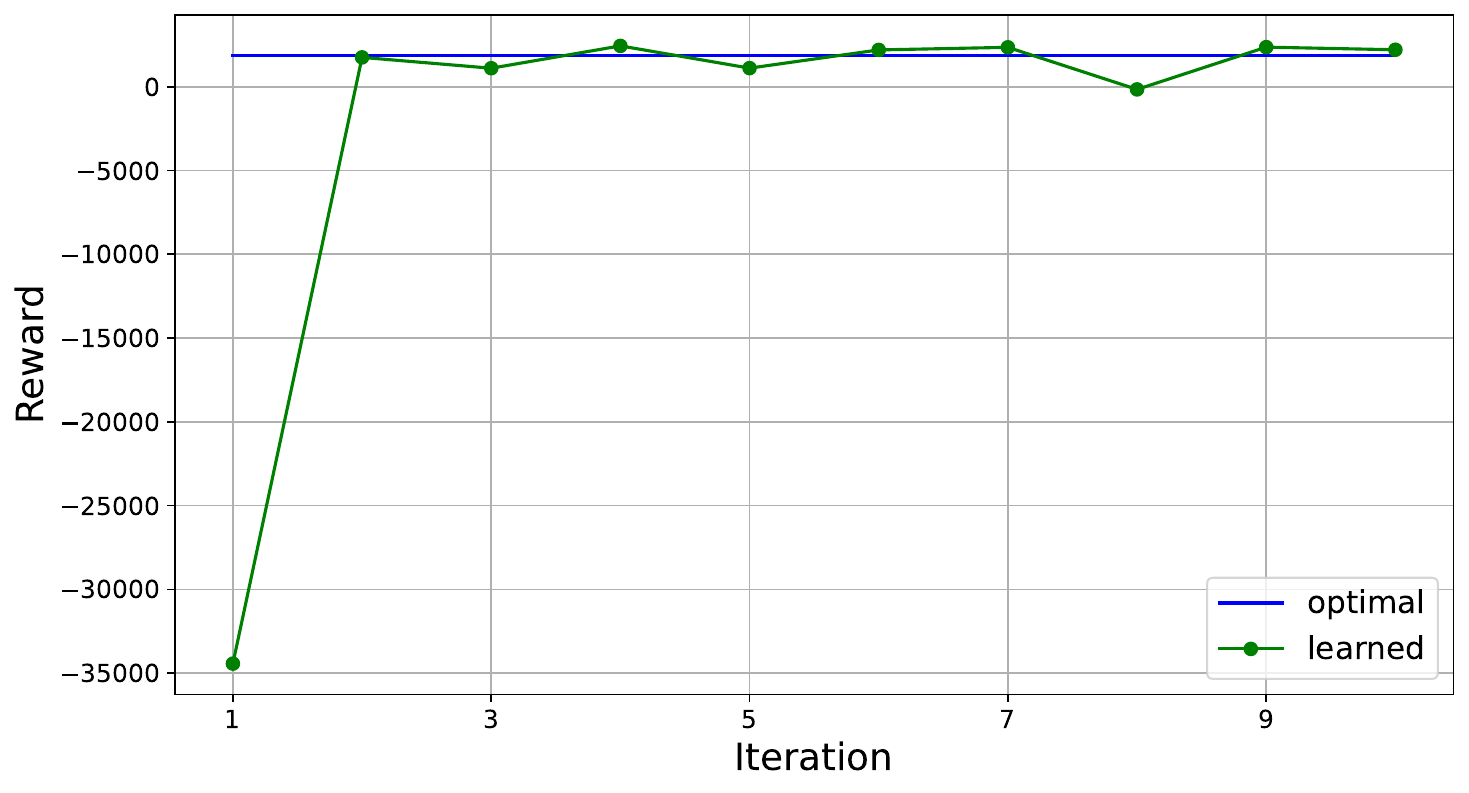}
         \caption{Obtained rewards on August 31st}
         \label{fig:rews243}
     \end{subfigure}
        \caption{Comparison of demand response provision obtained with true and learned reward evaluated with true reward.}
        \label{fig:four graphs}
\end{figure*}

\begin{table}[tb]
    \centering
    \caption{The average differences in demand response provision obtained with the true reward and with the learned reward for two days with different usage of controllable devices (April 2nd = low usage; August 31st = high usage).}
    \begin{tabular}{rrr}
       & April 2nd & August 31st \\
    \toprule
    MAE&0.0&0.12\\
    \midrule
    MSE&0.0&0.09\\
    \midrule
    Pearson&1.0&0.77\\
    \bottomrule

    \end{tabular}
    \label{tb:metricsS}
\end{table}

\subsection{Adaptability to Different Days}
The model is trained for 3500 episodes on the entire training set to determine the potential of the approach to approximate the reward on different consumption patterns. Testing is then done on 31 days from the test set and the results are presented in Table \ref{tb:generalmetrix}. The average (and median) MAE and mean squared error (MSE) on all tested days of 0.144 (0.159) and 0.065 (0.061) are comparable to the results for August 31 in Table \ref{tb:metricsS}, which leads to a conclusion of satisfactory model performance for different days. Among other things, dependency on the number of active controllable appliances on a certain day, resulted in the Pearson correlation coefficient for demand response provision based on the learned and optimal policy of 0.589 on the worst and 0.958 on the best tested day.

A more detailed view is provided in Fig. \ref{fig:bestgener}, where appliance schedules for the first twelve hours of August 8th under the optimal policy (Fig. \ref{fig:optimpolicy}) and learned policy (Fig. \ref{fig:learnedpolicy}) are compared. The values in these Figs. are in the interval $[-1, 1]$ for all appliances aside from the AC, where 1 means that an incoming request to start the appliance operation was ignored, 0 that it was accepted, and -1 that the device was started even without a prior request. The AC can take values from completely reduced consumption (0) to no reduction at all (1). Mostly, the agent's actions with both policies are the same on the observed day. The difference can be seen, e.g. in the second 15-minute period (red at the bottom) when the EV is charged with the optimal policy and is not charged with the learned one. Furthermore, there should have been some reductions of the AC consumption (light blue shades) based on the optimal policy, which did not happen with the learned one. Although a satisfactory approximation of the reward function can be obtained for certain days or parts of the day, the insufficient complexity of the presented model makes it impossible to achieve extremely high accuracy (e.g. less than 5\% of errors for all tested days). 

\begin{table}[tb]
    \centering
    \caption{MAEs, MSEs and Pearson correlation between demand provision with optimal and learned behaviours over the whole test set}

    \begin{tabular}{rrrrr}
       & Average & Minimum & Maximum & Median \\
    \toprule
    MAE&0.144&0.021&0.277&0.159\\
    \midrule
    MSE&0.065&0.020&0.167&0.061\\
    \midrule
    Pearson&-&0.589&0.958&0.843\\
    \bottomrule

    \end{tabular}
    \label{tb:generalmetrix}
\end{table}

\begin{figure*}[tb]
     \centering
     \begin{subfigure}[b]{0.475\textwidth}
         \centering
         \includegraphics[width=\textwidth]{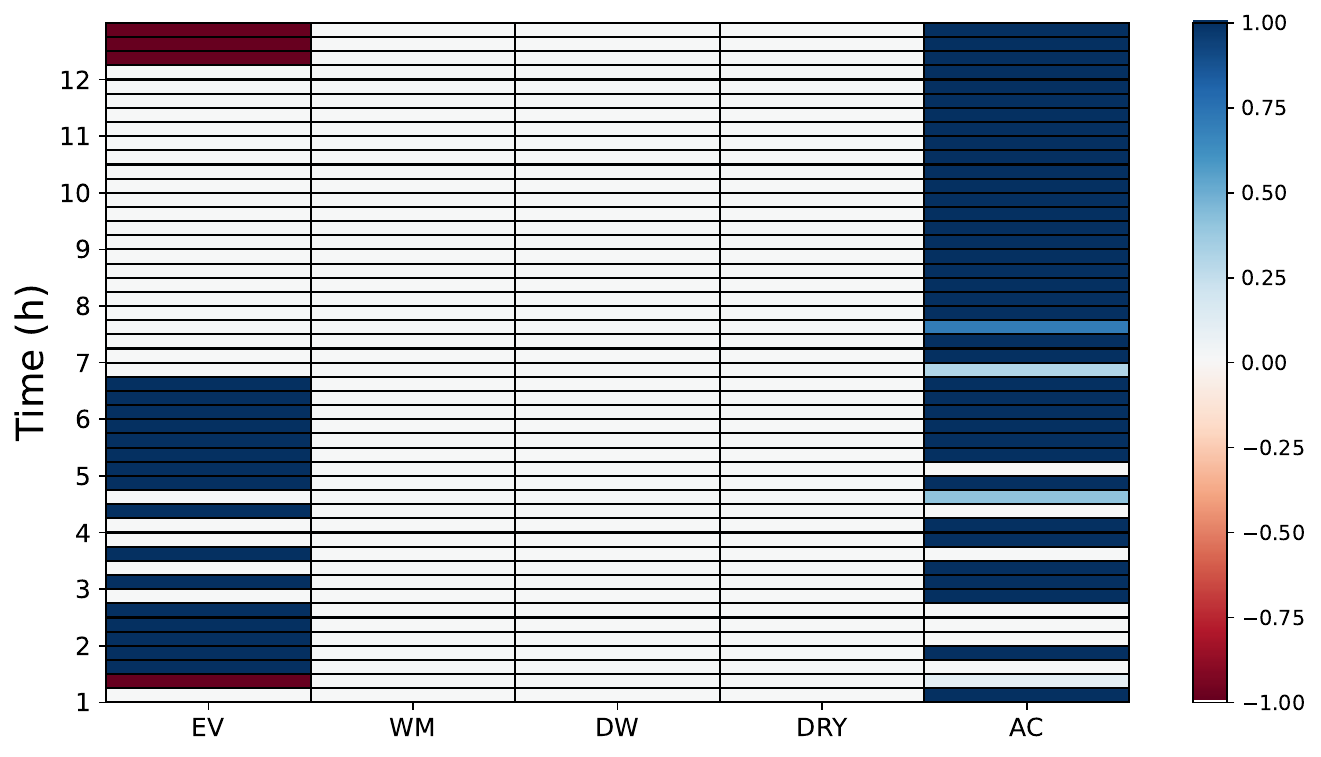}
         \caption{Optimal policy}
         \label{fig:optimpolicy}
     \end{subfigure}
     \hfill
     \begin{subfigure}[b]{0.475\textwidth}
         \centering
         \includegraphics[width=0.975\textwidth]{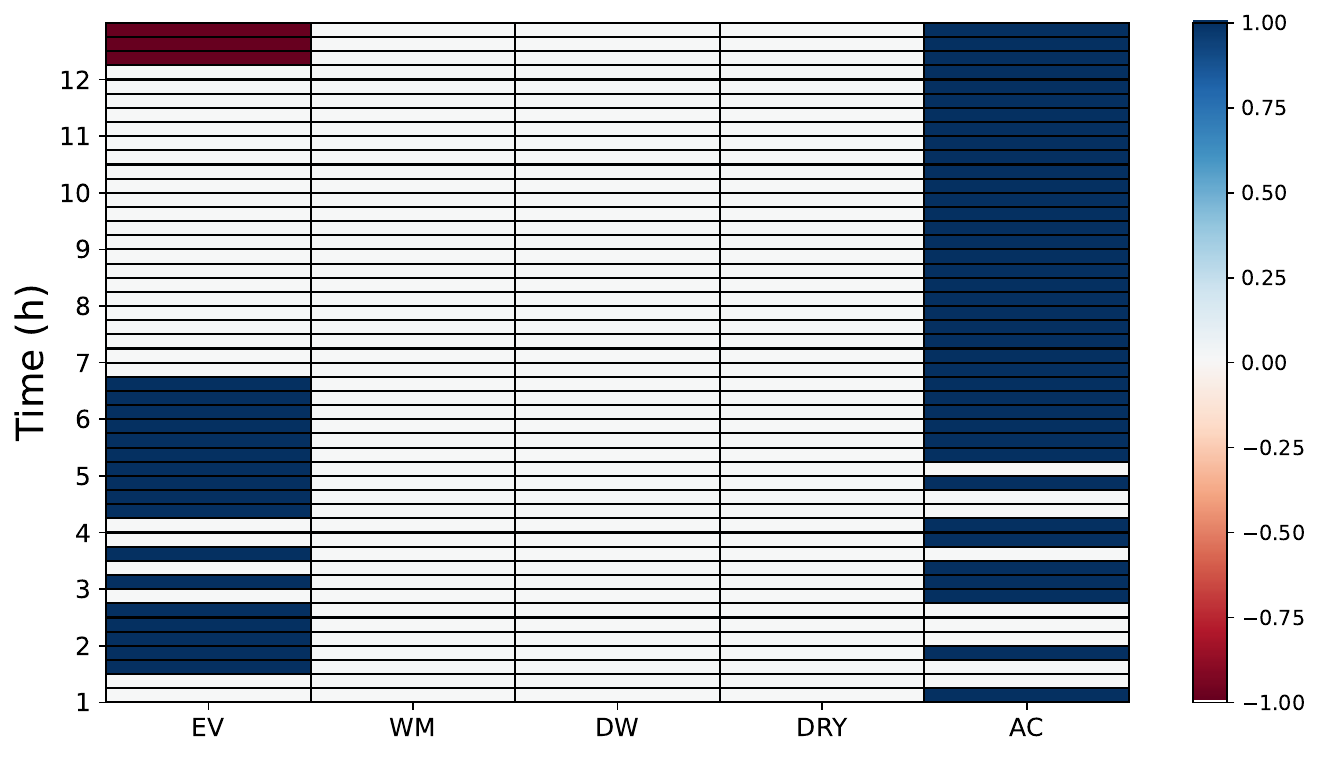}
         \caption{Learned policy}
         \label{fig:learnedpolicy}
     \end{subfigure}

        \caption{Comparison of the devices' schedule with optimal and learned policy tested on August 8th}
        \label{fig:bestgener}
\end{figure*}

\subsection{Adaptability to Different Users}
Concerning the adaptability to different households, five types of households with different underlying rewards were created that guide their scheduling decisions (actions). True rewards are assumed to vary in the way service provision revenue is modeled (either only the reductions or changes in both directions are rewarded) and in the way discomfort is modeled (absolute value or quadratic function). For four of them, optimizing coefficients $\alpha_i$ could lead to the exact true underlying reward of a particular household, since these true rewards are modeled to be equal to basis functions used in the IRL problem\footnote{One basis function is chosen for one household.}. However, one reward (household 5 in Fig. \ref{fig:householddist}) is modeled differently to investigate how the reward could adapt to a different environment. That reward is modeled as in \eqref{eq. reww2}, however, with different dissatisfaction ($w_{ac}, w_m$) coefficients. The resulting MAEs during the test period for all observed households are given in Fig. \ref{fig:householddist}. It is observed that for all households, at least 75\% of MAE falls below 0.20. Households 1 and 5 show a higher dispersity of errors (longer boxes) compared to the remaining households. Furthermore, distributions of MAEs in households 1, 2 and 4 are slightly positively skewed (range of MAEs higher than the median is wider than of those with MAEs lower than the median), while in households 3 and 5 a negative skew can be seen (more dispersed lower range of MAEs). The smallest errors, which are also the least scattered, occur in households 3 and 4, which in the background have a reward with an absolute value for discomfort and payment for changes of consumption in both directions (house 3) and only for reductions (house 4). Household 5, with a different underlying reward function that generated demand response consumption data, generally shows comparable MAEs to the remaining houses, but the total size of the interval within which the errors are located is much larger (the errors of the best and worst day in terms of MAE differ by approximately 30\%).

\begin{figure}[tb]
    \centering
    \includegraphics[width=1\linewidth]{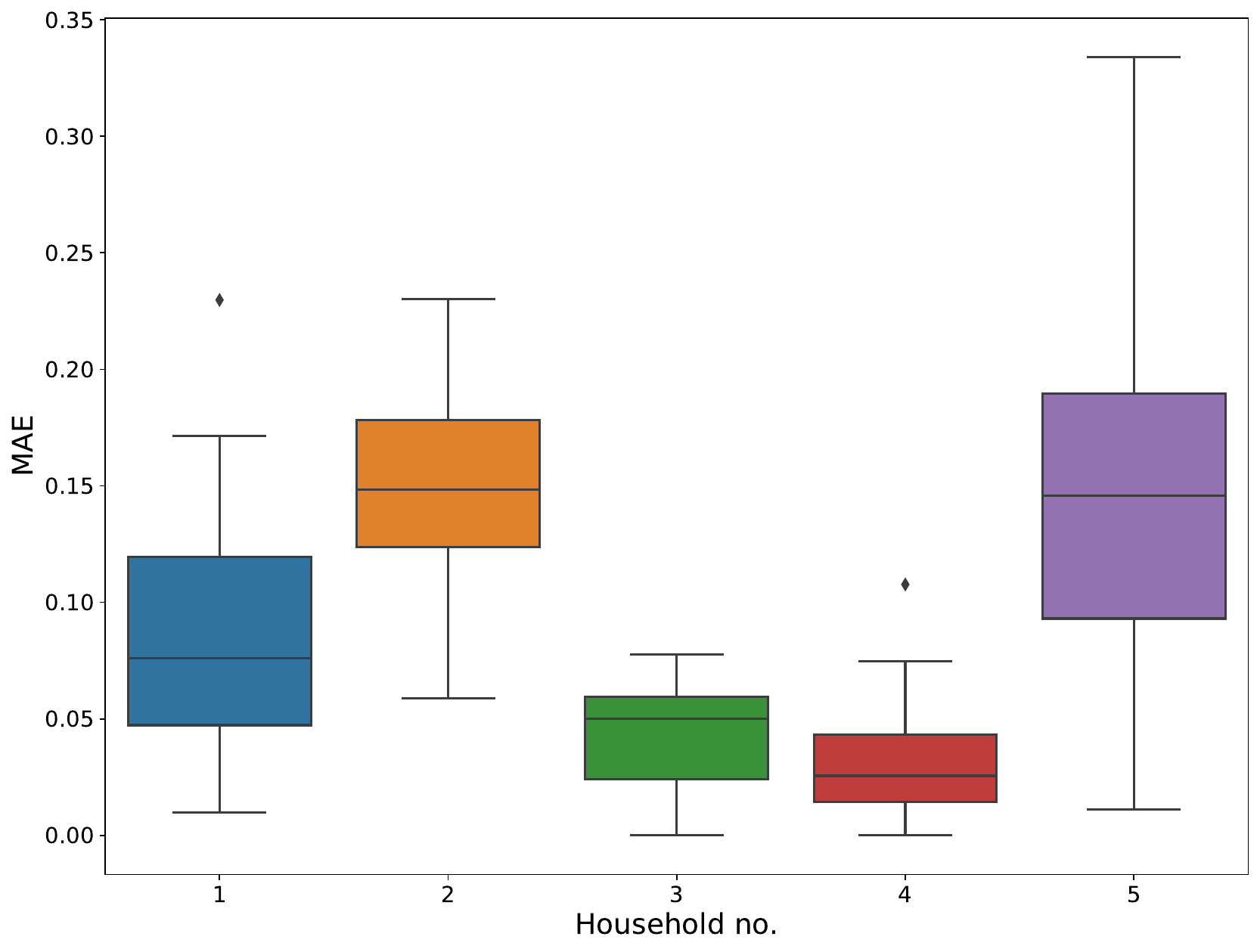}
    \caption{Distribution of MAEs between optimal and learned demand response provision for 5 households over the test set}
    \label{fig:householddist}
\end{figure}





\section{Conclusion}
To facilitate the provision of residential demand response, it is necessary, in addition to development of the infrastructure, to motivate the end users to participate. Most of them will decide to participate solely if the need for their engagement in controlling the household appliances is not high and if their comfort remains (almost) intact. To make this possible, an inverse reinforcement learning model for learning the reward function of a household that provides demand response is presented in this paper. Through learning the reward function from the historic data, unnecessary user engagement in decision-making process is avoided and their comfort is kept unharmed. 

The presented model showed satisfactory results of recreating the user-desired reward function. It is showed that the obtained reward function can generalize over different daily consumptions, however, the more complex the consumption (in terms of the number of active appliances) the harder it is to recreate the reward function. The model proved to be successful even when applied to different households with different consumption habits and desires.

Considering the complexity of the demand response problem, the presented model is too simple to significantly reduce errors in recreating the reward function. Assuming a family of functions for the reward function at the outset greatly reduces the solution space. Furthermore, the need to define the basis functions as a kind of input data also limits the solution space, and at the same time has a significant impact on the final result. In future work, it will be compared how more advanced techniques of inverse reinforcement learning, which do not restrict the reward function family of functions or learning the basis functions instead of assuming them known, may improve reward learning process and demand response provision. Lastly, it is important to emphasize that the inherent problem of trying to learn the objective/reward function in the demand response problem in this way is the impossibility of objectively defining a satisfactory result. For a certain household, the fulfillment of only half of its wishes may be enough, while on the other hand, another household could expect a perfect performance in order to find the model's work satisfactory. In future work, efforts will be made to define metrics in order to enable the most objective evaluation of the model's performance in this research area.

\end{document}